\documentclass[letterpaper, 10 pt, conference]{ieeeconf}  

\IEEEoverridecommandlockouts                              

\usepackage{amsmath, amssymb, amsfonts}
\usepackage{graphics} 
\usepackage{graphicx} %
\usepackage{stfloats}
\usepackage{float} %
\usepackage{subfigure} %
\usepackage{threeparttable}
\usepackage{tabularx}
\usepackage{multirow}
\usepackage{makecell}  
\usepackage{array}
\usepackage{algorithm}
\usepackage{algpseudocode}
\usepackage[normalem]{ulem}
\usepackage[colorlinks,linkcolor=red]{hyperref}
\usepackage{cite}
\usepackage{balance}

\usepackage{amsmath}

\usepackage{color}
\definecolor{b}{rgb}{0,0,1}
\makeatletter
\newcommand*{\new}{\@ifnextchar\bgroup{\new@}{\color{b}}}
\newcommand*{\new@}[1]{{\textcolor{b}{#1}}}

\title{\LARGE \bf
GAP-RL: Grasps As Points for RL Towards Dynamic Object Grasping
}

\author{
Pengwei Xie$^{1,*}$, 
Siang Chen$^{1,2,*}$, 
Qianrun Chen$^{5}$, 
Wei Tang$^{3}$, 
Dingchang Hu$^{1}$, 
Yixiang Dai$^{1}$, \\
Rui Chen$^{4,\dagger}$, 
Guijin Wang$^{1,2,\dagger}$%
    \thanks{$^{1}$Department of Electronic Engineering, Tsinghua University, Beijing 100084, China.}%
    \thanks{$^{2}$Shanghai AI Laboratory, Shanghai 200232, China.}
    \thanks{$^{3}$Department of Electronic Engineering, Shenzhen International Graduate School, Tsinghua University, Shenzhen 518071, China.}
    \thanks{$^{4}$Department of Mechanical Engineering, Tsinghua University, Beijing 100084, China.}
    \thanks{$^{5}$Risk, Reliability, and Resilience Engineering, Engineering school, Vanderbilt University, TN 37212, USA.}
    \thanks{$^{*}$Equal Contribution.}
    \thanks{$^{\dagger}$Corresponding Author: {\tt chenruithu@mail.tsinghua.edu.cn,\newline wangguijin@tsinghua.edu.cn.}}
}

\begin{document}

\maketitle

\begin{abstract}
Dynamic grasping of moving objects in complex, continuous motion scenarios remains challenging. Reinforcement Learning (RL) has been applied in various robotic manipulation tasks, benefiting from its closed-loop property. However, existing RL-based methods do not fully explore the potential for enhancing visual representations. In this letter, we propose a novel framework called Grasps As Points for RL (GAP-RL) to effectively and reliably grasp moving objects. By implementing a fast region-based grasp detector, we build a Grasp Encoder by transforming 6D grasp poses into Gaussian points and extracting grasp features as a higher-level abstraction than the original object point features. Additionally, we develop a Graspable Region Explorer for real-world deployment, which searches for consistent graspable regions, enabling smoother grasp generation and stable policy execution. To assess the performance fairly, we construct a simulated dynamic grasping benchmark involving objects with various complex motions. Experiment results demonstrate that our method effectively generalizes to novel objects and unseen dynamic motions compared to other baselines. Real-world experiments further validate the framework's sim-to-real transferability.

\end{abstract}

\section{INTRODUCTION}
Robotic grasping is a longstanding challenge in embodied AI research \cite{xiang2020sapien, fang2023anygrasp, christen2023handover, wang2024genh2r}. While significant progress has been made in vision-guided grasping of static objects, it remains an open challenge to grasp moving objects reliably. Developing strong capabilities for grasping dynamic objects has the potential to handle a wide range of applications, from factory assembly lines to seamless human-robot collaborative scenarios.



The most direct method for grasping moving objects is online grasp tracking, which tracks and approaches the closest grasp \cite{morrison2020learning}. However, this straightforward strategy requires consistently accurate grasp detection, resulting in jerky robot movements. Many studies instead use object tracking methods combined with specific motion planning techniques to track and grasp objects \cite{tuscher2021deep}. A notable limitation of these approaches is the requirement for the target object to remain stationary during the grasp execution, which restricts the system's ability to handle complex dynamic tasks. While some methods \cite{akinola2021dynamic, jia2023learning} can handle moving objects, they rely on object motion prediction and require careful tuning of critical hyperparameters. This complexity makes dynamic grasping difficult, particularly with novel object shapes and unpredictable motions.


Reinforcement Learning (RL) based methods have gained popularity due to their closed-loop property. GA-DDPG \cite{wang2022goal} processes the original object point cloud to learn an implicit representation, incorporating target grasp prediction as an auxiliary task. GAMMA \cite{zhang2023gamma} starts with scene-level grasp detection and represent the grasps using 9D features (3D translations and 6D rotations). EARL \cite{huang2023earl} uses pre-generated grasp proposals, converting them into multiple keypoints sampled from a simple gripper model for RL learning. However, these visual representations learned from limited objects potentially exhibit poor generalization to novel objects.

\begin{figure}[t]
\centering
    \includegraphics[width=8.5cm]{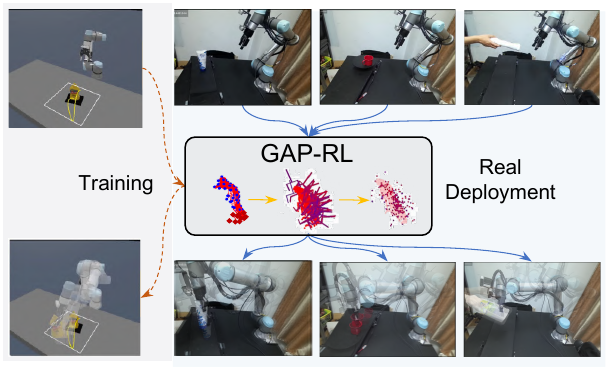}
    \caption{Our 3D RL framework is trained with various objects in simulation efficiently. After training, The RL policy can be deployed to a real robot and extends to various dynamic tasks effectively.}
    \label{fig:teaser} 
\vspace{-0.5cm} 
\end{figure}


In this work, we propose a novel RL-based framework for dynamic object grasping, termed as \textbf{GAP-RL}, representing \textbf{G}rasps \textbf{A}s \textbf{P}oints to assist the learning of the \textbf{RL} policy. As depicted in Fig. \ref{fig:teaser}, our framework is trained in simulation with various objects, and it can be effectively deployed on real robots to manage dynamic grasping tasks involving novel objects across various motion trajectories. The pipeline begins with the a Graspable Region Explorer to identify the consistent graspable regions. Subsequently, built upon Local Grasp (LoG) \cite{tang2024rethinking}, we implement a fast region-based grasp detector to generate high-quality grasps. To support the RL policy, each grasp is transformed into random Gaussian points, from which hierarchical grasp features are extracted. This innovative grasp-guided approach diverges from traditional visual representations such as point clouds and depth maps and is specifically designed to manage dynamic tasks involving novel objects across diverse scenarios. It effectively facilitates direct and efficient sim-to-real transfer.

To assess the performance, we construct a dynamic object grasping simulation benchmark in SAPIEN \cite{xiang2020sapien}. In this benchmark, the object undergoes diverse dynamic motions and exhibits varying speeds in both translation and rotation along the z-axis. Comprehensive evaluations in both simulation and real-world scenarios show that GAP-RL excels in accurately approaching and grasping novel moving objects, achieving high success rates across various motion modes.

Our main contributions are as follows:
\begin{itemize}
    \item We present GAP-RL, a novel RL framework for dynamic object grasping. GAP-RL represents grasps as points and extracts hierarchical features, enhancing generalization abilities across novel scenarios.
    \item We propose a Region Center Explorer to search for consistent graspable regions from consecutive frames, facilitating smoother grasp generation and more stable policy execution in the manipulation process.
    \item We develop a simulated dynamic grasping benchmark and then apply in real-world evaluations. The results illustrate that GAP-RL exhibits remarkable generalization abilities and robust sim-to-real transferability.
\end{itemize}

Our code and benchmark environment will be open-sourced to to promote the research of robot grasping.

\section{RELATED WORKS}

\subsection{Dynamic Grasping}   \label{related_dynamicgrasping}
Some previous works rely on object detection \cite{allen1993automated, chen2021deep} to track and grasp the object by pre-defined commands, limiting their adaptability to novel objects. Another line of work integrates object 6D pose tracking with pre-generated candidate grasps for the target object \cite{marturi2019dynamic, tuscher2021deep, akinola2021dynamic, jia2023learning}. However, 6D pose tracking may be time-consuming and usually requires objects' prior knowledge. Recent research has shifted from \textit{object tracking} to \textit{grasp tracking}. They are dedicated to planning temporally smooth \cite{fang2023anygrasp, zhang2023flexible} or semantically consistent \cite{liu2023target} grasps to avoid wavy and jerky robot motions. Differently, our method employs a grasp-guided RL policy to capture the moving object in a closed-loop system, requiring no additional estimation of the object's shape, pose, or movement states.

The integration of RL enables the acquisition of manipulation skills without the need for explicit motion planning. QT-Opt \cite{kalashnikov2018scalable} learns from large-scale data to grasp objects and respond dynamically to disturbances, but it may struggle with objects exhibiting more complex motions. GA-DDPG \cite{wang2022goal} processes segmented object points to learn manipulation policies, enhanced by goal-auxiliary tasks and expert data augmentation. Christen et al. \cite{christen2023learning} further refine the approach with a two-stage teacher-student framework, still taking the original object points as input. These works demonstrate the utilization of object points as input without explicit grasp guidance and require an extra instance segmentation model.

Recent advancements have shown the effectiveness of utilizing grasp detection results. GAMMA \cite{zhang2023gamma} represents grasps with 9D features (3D translations and 6D continuous rotation representations) and learns mobile manipulation policies with online grasp fusion. EARL \cite{huang2023earl} represents target grasp using four keypoints to capture both translation and rotation, employing a curriculum design that guides the policy to track and grasp the object when it becomes stationary. While these approaches integrate multiple external modules to support RL policy learning, the efficacy of their grasp representations for novel objects and diverse dynamic motions remains unverified. In contrast, our approach transforms grasps into random Gaussian points and extracts hierarchical geometric features, enabling robust generalization across diverse objects and motion patterns.

\subsection{Visual Representations for RL}
Visual representations for RL in robotic tasks have long been explored. Earlier works leverage 2D images to guide RL policy learning \cite{hsu2022visionbased, zhu2024point}, but recently, 3D point clouds have gained traction \cite{wang2022goal, xie2023part, huang2023earl, zhang2023gamma, wang2024genh2r}. 3D point clouds, which reveal spatial relationships within scenes, are ideally suited for tasks requiring intricate spatial reasoning \cite{ling2023efficacy, zhu2024point} and have demonstrated strong sim-to-real transferability. This is largely due to their focus on geometry rather than texture, simplifying the sim-to-real transition \cite{zhang2023close}. Despite the advantages, most approaches rely on PointNet \cite{qi2017pointnet} or PointNet++ \cite{qi2017pointnet++} to directly extract geometric features from segmented target point clouds. 

In this work, we consider detected grasps on the target object as a higher-level abstraction than the original object points for dynamic grasping tasks. Leveraging the proven efficiency of 3D RL policies for robotic manipulation, we represent grasps as combinations of random Gaussian points. This novel representation not only shows strong adaptability in simulation but also effectively extends to diverse real-world scenarios.

 
\section{Problem Statement}

We consider a setting where a robot approaches and grasps moving objects within a workspace. 
The dynamic setup assumes that objects are rigid bodies and can be manipulated by the typical two-finger gripper, without requiring specific characteristics related to the object's shape or motion states.
Moreover, the objects are allowed to move freely within the robot's operational area at random speeds. The system relies solely on egocentric depth images and essential robot state information to facilitate learning to control the end-effector in Cartesian space and adjust the gripper positions accordingly. The task is deemed successful if the robot can approach and securely grasp the object within a predefined time limit. 

In this setting, the dynamic object grasping problem can be modeled as a Partially Observable Markov Decision Process (POMDP), aiming to learn a policy $\pi:\mathcal{O}\rightarrow \mathcal{A}$. The observation $o \in \mathcal{O}$ includes visual features from a hand-centric camera, and the action $a \in \mathcal{A}$ indicates the robot target 3D Cartesian positions, 3D Euler angles and the gripper finger position.

\begin{figure*}[t]
\centering
    \includegraphics[width=0.95\textwidth]{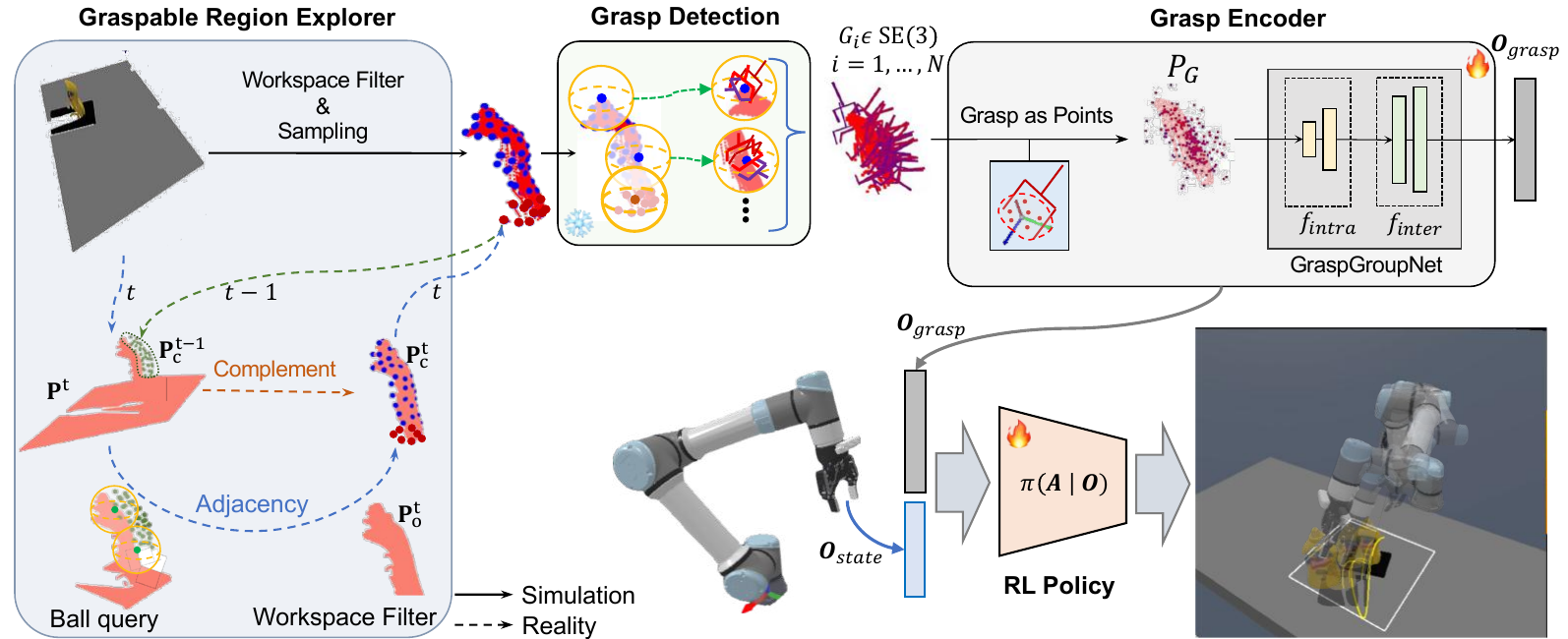}
    \caption{\textbf{GAP-RL Framework Overview}. In the simulation environment, GAP-RL inputs a hand-centric point cloud and explores graspable regions by workspace filtering and sampling region centers. These region centers are then used to crop spherical neighborhood points and detect dense grasp results. We then represent these grasps as Gaussian points and extract features $\mathbf{O}_{grasp}$ from GraspGroupNet (GGN). $\mathbf{O}_{grasp}$ along with agent states $\mathbf{O}_{state}$ are combined for learning $\pi(\mathcal{A}|\mathcal{O})$. In real deployment, to reduce the impact of motion blur, we design a Region Center Explorer to extract graspable regions of the target object adaptively.}
    \label{fig:Frame} 
\vspace{-0.2cm} 
\end{figure*}

\section{METHOD}  \label{method}
\subsection{GAP-RL Framework}

As depicted in Fig. \ref{fig:Frame}. we develop a grasp-guided RL framework for dynamic object grasping tasks. We employ a hand-centric RGBD camera to mitigate occlusion problems and adapt to real applications easily. 

Initially, the agent takes in the depth image and transforms the depth pixels to 3D points in the world frame, using the calibrated camera's intrinsic and extrinsic parameters. Then, it explores the graspable regions within the workspace and samples grasp center points from these regions. Local region points are obtained by a ball query strategy \cite{qi2017pointnet++}. Next, we leverage the Local Grasp model \cite{tang2024rethinking} to generate multiple 6D grasps around the target object online. To ensure a robust grasp representation, we transform grasps into random Gaussian points and aggregate all the grasp points. Subsequently, a hierarchical grasp point network named GraspGroupNet (GGN) is utilized to group and condense the features. Finally, the geometric features are fed into the RL algorithm with robot states to predict actions. 

Our unified framework adaptively explores graspable regions of moving targets and generates high-quality grasps to guide the RL policy. This framework does not rely on pre-assumptions about the object's shape or motion state, making it straightforward and robust to implement in real-world scenarios.


\subsection{Graspable Region Explorer}
This module is designed to efficiently track graspable regions on the target object, facilitating high-quality grasp detection. Previous research often relies on segmentation \cite{wang2022goal} or point tracking \cite{doersch2023tapir, vecerik2024robotap} from consecutive RGB frames, requiring additional data modalities and model processing. In contrast, our module leverages grasp detection results with only depth input, making it more straightforward and fully compatible with our grasp detection process.


In simulation, region centers can be directly sampled by cropping the object using a workspace filter. However, in real scenarios, motion blur can introduce noise, potentially degrading the process. Since the object moves continuously without abrupt changes in position, region centers can be efficiently sampled by considering the adjacent graspable regions. In real experiments, as outlined in Algorithm \ref{Alg.GraspExplore}, we propose a region center explorer algorithm that explores adjacent graspable regions and samples grasp centers for grasp detection. To prevent losing graspable regions during motion, we retain several complementary centers from the points filtered by the workspace.

The algorithm offers two main advantages over direct sampling from the points filtered by the workspace. First, motion blur introduces severe noise into the point cloud, and distant noisy points may not be effectively filtered. Our algorithm focuses on sampling region centers from neighboring graspable regions of the target object, which are less affected by dynamic noise. Second, the temporal consistency of the target object points makes our algorithm suitable for complex dynamic settings in simulations and enables seamless adaptation to the 3D workspace.

\newcommand{\pushcode}[1][1]{\hskip\dimexpr#1\algorithmicindent\relax}
\begin{algorithm}[tbp]
\definecolor{codeblue}{rgb}{0.25,0.5,0.7}
\caption{Region Center Explorer}
\label{Alg.GraspExplore}
\textbf{Require:}\\
$T$ - episode steps, $ws$ - predefined workspace, 
\\ $\mathbf{P}_c^{t-1}$ - region centers in the previous frame, 
\\ $\mathbf{P}^t$ - points in the current frame,
\\ $N_o$ - number of the tracked centers from the target points.
\\ $N_c$ - number of the complement centers. \\
\textbf{Pseudocode:} 
\begin{algorithmic}[1]
\For{$t=1\rightarrow T$}
\State $\{\mathbf{P}^t_{i}|i=1,...,K\} \gets \text{ball\_query}(\mathbf{P}^t,\text{centers=}\mathbf{P}_c^{t-1})$
\Statex \quad\ {\color{codeblue}\# Crop regional points.} 
\State $\mathbf{P}_o^t \gets$ Filter(Concat($\mathbf{P}^t_{i}), ws$)
\Statex \quad\ {\color{codeblue}\# Aggregating regional points and filtering.} 
\State $\mathbf{P}_{target}^{t} \gets$ FPS($\mathbf{P}_o^t, N_o$)
\Statex \quad\ {\color{codeblue}\# Farthest Point Sampling (FPS) from target points.}
\State $\mathbf{P}_{comple}^{t} \gets$ FPS(Filter($\mathbf{P}^t, ws$), $N_c$)
\Statex \quad\ {\color{codeblue}\# Retain several complement centers.} 
\State $\mathbf{P}_c^{t} \gets \text{Concat}(\mathbf{P}_{target}^{t}, \mathbf{P}_{comple}^{t})$
\Statex \quad\ {\color{codeblue}\# Get all region centers.} 
\State $\hat{\mathbf{P}}_c^{t} \gets \text{Filter}(\mathbf{P}_c^{t}, d_{th})$
\Statex \quad\ {\color{codeblue}\# Filtering the centers above the distance threshold.}
\EndFor
\end{algorithmic}
\end{algorithm}

\subsection{Grasp Detection}

In this module, we utilize Local Grasp (LoG) \cite{tang2024rethinking} to detect grasps $\mathbf{\hat{G}}$ from graspable region centers $\hat{\mathbf{P}}_c$ due to its high efficiency and strong performance for novel objects.

We adopt the grasp representation in \cite{tang2024rethinking, chen2023hggd}. Consider a set of regional points, which are centered at $(x_p,y_p,z_p)$ in the camera frame. Within the region, a region-level grasp $\mathbf{\hat{g}}_{j} = (x_p+\Delta x, y_p+\Delta y, z_p+\Delta z, \theta, \gamma, \beta, w_g) \in \mathbf{\hat{G}}$ is centered at $(x_p,y_p,z_p)$.
Here, $(\Delta x, \Delta y, \Delta z)\in \mathbb{R}^3$ represent the offset between the center of the grasp and the region center. $(\theta,\gamma,\beta)\in [-\frac{\pi}{2}, \frac{\pi}{2}]$ are grasp Euler angles and $w_g$ denotes the grasp width. More details can be referred to \cite{tang2024rethinking}.

To accommodate dynamic settings, specific adjustments and improvements have been made to its configurations. Originally, LoG requires an additional guidance module to locate graspable areas within the scene. Now we explore region centers $\hat{\mathbf{P}}_c \in \mathbb{R}^{K\times 3}$ using Algorithm \ref{Alg.GraspExplore}, followed by the application of the ball query method to extract region points $R_i \in \mathbb{R}^{M \times 3}, i=1,..., K$. These points are then transformed from the camera frame to the region frame, centering the points at the region centers. This normalization is crucial as it mitigates the impact of varying distances between the agent and the object during the dynamic process. The grasp poses $\mathbf{\hat{G}}=\{\mathbf{\hat{g}}_{j},j=1,...,N\}$ are predicted by LoG denoted as $\Phi$. The entire process can be formulated as $\mathbf{\hat{G}} = \mathcal{T}^{R}_{EE}(\Phi(\mathcal{T}^{C}_{R}(R_i | i=1,..., K)))$,
where $\mathcal{T}^{C}_{R}$, $\mathcal{T}^{R}_{EE}$ denote the transformations. Notably, $\mathbf{\hat{G}}$ is represented in the end-effector frame, a method proven to enhance sample efficiency in robotic manipulation tasks \cite{liu2022frame}.

In our experiments, the adjusted grasp detector detects grasps from graspable regions quickly, achieving an inference frequency of approximately 20Hz. This rapid response allows our pipeline to adapt immediately to changes.



\subsection{Grasp Encoder} \label{GraspEncoder}
After obtaining $\mathbf{\hat{G}}$, each grasp is represented as points. Then we aggregate the points and extract the hierarchical features $\mathbf{O}_{grasp}$ as an abstract guidance.

For single grasp, as discussed in Section \ref{related_dynamicgrasping}, previous works \cite{qin2020s4g, liu2023target, zhang2023gamma} introduce 9D features to represent pose in a straightforward manner. And they train a simple network like PointNet to capture implicit relationships in $\mathrm{SE(3)}$ space among various grasp poses. Additionally, by transforming each grasp to a fixed set of keypoints from an abstract gripper model \cite{wang2022goal, huang2023earl}, relationships among 6D poses can be transformed into explicit distances among 3D point clouds. However, this gripper-specific fixed-point representation might result in network overfitting to specific gripper geometries.

To overcome this, inspired by Neural Pose Descriptor Fields \cite{simeonov2022neural}, we transform each grasp into a randomly generated set of points from a Gaussian distribution in each episode. Such representation can capture the relative relationships among different poses well, making it compatible with 3D RL methods.




\subsubsection{Grasp as Points}
We start by initializing a set of Gaussian points: 
$\chi = \{\mathbf{x}_i\}_{i=1}^L \sim \mathcal{N}(\mathbf{0}, \sigma^2 \mathbf{I}),$ where $\chi\in \mathbb{R}^{L\times 3}$ is the initial points and the scale of $\chi$ is parameterized by the gripper width as $\sigma = w_g / 6$, which encapsulate most points while no restrictions for the gripper shape. Built upon this, any grasp pose can be represented as a transformed point set $\mathcal{T}\chi_h$, where $\mathcal{T}$ is the corresponding transformation and $\chi_h$ is the homogeneous coordinates of $\chi$. 


Obtaining the grasp poses $\mathbf{\hat{G}}$, we apply the corresponding transformations $\hat{\mathcal{T}}$ to the initial points set. The grasp points are then computed as: $\mathbf{\hat{P}_\mathbf{G}}=\{\hat{\mathcal{T}_j}\chi_{h}, j=1,...,N\}.$

\subsubsection{Grasp Points Encoder}
To extract grasp features from the grasp points $\mathbf{\hat{P}_\mathbf{G}}$, previous studies \cite{zhang2023gamma, huang2023earl, liu2023target} utilize typical point encoders (PointNet, PointNet++, etc.) to process multiple grasps directly. However, this approach may potentially neglect intra-grasp details, such as the orientations of the grasp pose, and could result in degraded features when directly applied to $\mathbf{\hat{P}_\mathbf{G}}$. In response to this concern, GraspGroupNet (GGN) is developed to focus on capturing intra-grasp details from $\hat{\mathcal{T}_j}\chi_h$ for each grasp first. Subsequently, it aggregates these intra-grasp features to further extract and analyze inter-grasp relationships, enhancing the overall grasp representation.

Specifically, two shared MLP (multi-layer perceptron) layers are utilized to encode the transformed points $\mathbf{\hat{P}_\mathbf{G}}$. Then, a max pooling layer is applied within each grasp point group to derive invariant intra-grasp embeddings. These embeddings are subsequently processed through another two MLP layers followed by a grasp-wise max pooling layer, which condenses them into the compact grasp features denoted as $\mathbf{O}_{grasp}$.

\subsection{Grasp-Guided RL Policy Learning}
The RL policy is trained using the Soft Actor-Critic (SAC) algorithm \cite{haarnoja2018soft}. The observations include compact grasp features $\mathbf{O}_{grasp}$ and robot state features $\mathbf{O}_{state}$. $\mathbf{O}_{state}$ specifically captures robot arm's end-effector pose and the current positions of the gripper fingers. Our approach enables the policy to output a 6-DoF residual adjustment for the robot arm's Cartesian pose, along with a 1-DoF action to control the gripper's opening and closing.

\subsubsection{Reward Shaping}
The reward function is structured to guide the agent through the three stages of the task: approaching, grasping, and lifting the moving object. It comprises four components: 1) An approaching term that minimizes the distance to the nearest annotated grasp pose (target grasp, see \ref{goalaux}), aligning the agent with the optimal grasp configuration. 2) A grasping term that promotes secure object grasping. 3) A lifting term that rewards successful object elevation, indicating task completion. 4) A grasp visibility term that ensures detected grasps within the egocentric camera's field of view. While these components are standard in grasping tasks, we introduce a novel distance metric that enhances performance:

\begin{equation}
\label{eq.grasp_dist}
d(\mathbf{\hat{G}}, \mathbf{g}_{EE}) = \mathop{\min}_{j=1,...,N} ||\mathbf{p}_{EE}-\mathbf{\hat{p}_j}||^2 + (1-|\mathbf{\hat{q}_j}\cdot \mathbf{q}_{EE}|).
\end{equation}
Here, $(\mathbf{\hat{p}_j}, \mathbf{\hat{q}_j})$ represents the position and quaternion of the grasp pose $\mathbf{\hat{g}}_j$, and $(\mathbf{p}_{EE}, \mathbf{q}_{EE})$ denotes the end-effector's pose $\mathbf{g}_{EE}$. 

The metric comprises two components: the position Euclidean distance and a rotational distance from Huynh et al. \cite{huynh2009metrics}. This approach is preferred over conventional object-to-end-effector distance due to its alignment with our grasp observations and auxiliary goal task.

\subsubsection{Auxiliary Refining} \label{goalaux}
Self-supervised tasks have been proven effective for RL policy learning by providing auxiliary losses \cite{shelhamer2016loss}. In dynamic grasping tasks, relying solely on online grasp observations and robot state is insufficient for enabling the policy to adeptly interact with the objects. 
A key motivation is enabling the agent to effectively engage with the environment across different stages, potentially improving performance.
To achieve this, we leverage the reward as a proxy task for predicting the current stage, which can be expressed as $\mathbf{\hat{s}}_{pred} = f(\mathbf{O}_{grasp}, \mathbf{O}_{state}), \mathbf{\hat{s}}_{pred}\in \{0, 1\}^3.$
Here we model the task as a multi-label binary classification task and utilize the binary cross-entropy loss as $L_{target}$. 

Moreover, another difficult problem is encouraging the agent to grasp the object in a proper pose. Inspired by Wang et al. \cite{wang2022goal}, we collect grasp annotations from GraspNet-1billion Dataset \cite{fang2020graspnet} as target grasps $\mathbf{G}^*$. 

And an auxiliary grasp goal prediction task is employed to guide the agent with the nearest target, which can be expressed as $\mathbf{\hat{g}}_{pred} = g(\mathbf{O}_{grasp}, \mathbf{O}_{state}), \mathbf{\hat{g}}_{pred} \in \mathbb{R}^7, ||\mathbf{\hat{q}}_{pred}|| = 1,$
where $\mathbf{\hat{g}}_{pred}$ is the predicted target pose and $\mathbf{\hat{q}}_{pred}$ is the related quaternion. And the auxiliary loss, denoted as $L_{goal}$, is computed using $d(\mathbf{G}^*, \mathbf{\hat{g}}_{pred})$ from Equation \ref{eq.grasp_dist}.


The integrated network, combining the Grasp Encoder and the actor-critic networks, is updated using the original SAC loss $L_{SAC}$ with two additional terms $Loss = L_{SAC} + (\alpha * L_{target} + \beta * L_{goal}) * \gamma^\frac{t}{20000},$
where $L_{target}$ and $L_{goal}$ primarily refine the Grasp Encoder, which has been proved effective for RL-based robotic grasping tasks \cite{shelhamer2016loss, wang2022goal}. Here, $\alpha$ and $\beta$ are weighting factor, $\gamma$ is the discount factor, and $t$ represents the total timesteps in the learning process.



\subsection{Implementation Details}
In our experiments, we set $M$ to 512, following \cite{tang2024rethinking}, and $K$, $N$, $N_o$, $N_c$ and $L$ to 64, 40, 48, 16, and 20, respectively, to achieve a balance between performance and efficiency. The parameters $\alpha, \beta, \gamma$ are set to 1, 1 and 0.98, respectively. More details about our model architecture can be found in the supplementary materials. During training, the Grasp Encoder and RL networks are trained from scratch over two million steps (20,000 episodes, 100 steps per episode), which takes approximately 6 hours on a desktop with an i9-13900K CPU and an NVIDIA RTX 4090 GPU.

\section{EXPERIMENT}
We train the framework in simulation. To thoroughly evaluate the performance, we conduct experiments both in simulation and implement on the real robot platform.

\subsection{Simulation Setups} \label{Sim_Setup}

\begin{figure}[t]
\centering
    \includegraphics[width=8cm]{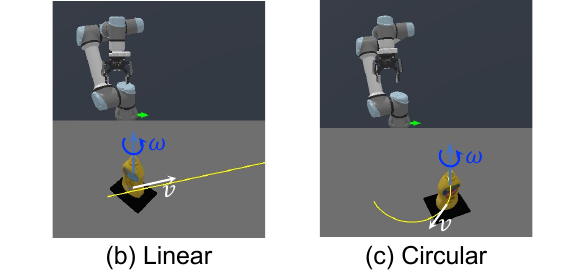}
    \caption{The dynamic settings in the simulation environment.}
    \label{fig:DySet} 
\vspace{-0.5cm} 
\end{figure}

As illustrated in Fig. \ref{fig:DySet}, we construct different dynamic scenarios in the Sapien simulator \cite{xiang2020sapien} to assess performance. Details about the range of all parameters are provided in the supplementary materials. In these scenarios, the target object moves in different motion modes.

\begin{itemize}
  \item \textit{Rotation}: The object is randomly placed in the workspace.
  \item \textit{Line}: The object moves along a straight line from one end to the other at a random speed.
  \item \textit{Circle}: The object follows a circular trajectory with a randomly positioned center and variable radius within the workspace.
  \item \textit{Random}: The object traces a randomly generated Bezier curve at a random speed, confined within the workspace.
\end{itemize}

To increase the challenge of the scene, we make the object rotate around the \textit{z}-axis at a random angular velocity $\omega \in [-10\ ^\circ/s, 10\ ^\circ/s]$ while following a specified trajectory. Note that as the object's motion progresses from \textit{Rotation} to \textit{Random}, the difficulty escalates due to increasing degrees of freedom. Training on the most complex mode, \textit{Random}, equips the agent to handle diverse, dynamic motions and successfully grasp the target.

To evaluate the effectiveness of GAP-RL, we compare the performance of the following methods.
\begin{itemize}
  \item \textbf{Heuristic Planning}: Using the tracking strategy proposed in \cite{morrison2020learning}, the agent continuously tracks the nearest grasp until reaching a threshold. Subsequently, the agent moves the gripper forward to align with the grasp pose, closes the gripper, and lifts the object.
  \item \textbf{GA-DDPG} \cite{wang2022goal}: This method utilizes the object point representations with goal-auxiliary tasks, while replacing the vanilla DDPG with SAC to ensure a fair comparison.
  \item \textbf{GAMMA} \cite{zhang2023gamma}: 9D grasp features are used to represent a single grasp, combining 3D translations with continuous 6D rotations \cite{zhou2019continuity}. And grasp features are extracted using the same grasping pose encoder in \cite{zhang2023gamma}.
  \item \textbf{EARL} \cite{huang2023earl}: A fixed set of keypoints is used to represent a single grasp, replacing the random Gaussian points $\mathbf{\hat{P}_\mathbf{G}}$ in our framework.
  \item \textbf{GAP-RL}: Our framework as discussed in Section \ref{method}.
\end{itemize}

Note that we have reproduced the baselines \cite{wang2022goal, zhang2023gamma, huang2023earl}, retaining their visual representations while keeping all other settings consistent to ensure a fair comparison. Apart from Heuristic Planning, we employ these visual representations to facilitate the RL policy learning.

\subsection{Performance Evaluation} \label{Sim_Exp}

\begin{table*}[t]
\caption{Results on different simulation settings. The results are averaged over 5 random seeds.}
\label{tab:main_exp}
\centering
\small
\renewcommand\arraystretch{1.3}
\resizebox{0.9\textwidth}{!}{
\begin{threeparttable}
\begin{tabular}{c|cc|cc|cc}
\hline
\multicolumn{1}{c|}{\multirow{2}{*}{}} & \multicolumn{2}{c|}{\textbf{Seen Split}}  & \multicolumn{2}{c|}{\textbf{Unseen Split}} & \multicolumn{2}{c}{\textbf{Novel Split}}            
\\
\multicolumn{1}{c|}{}  & \textit{Seen Traj.} & \textit{Novel Traj.} & \textit{Seen Traj.} & \textit{Novel Traj.} & \textit{Seen Traj.} & \textit{Novel Traj.}
\\ \hline


Heuristic Planning & 0.246 & 0.312 & 0.225 & 0.333 & 0.200 & 0.210
\\

GAMMA$^\dagger$ \cite{zhang2023gamma} & $0.235\pm0.133$ & $0.207\pm0.120$ & $0.267\pm0.126$ & $0.219\pm0.112$& $0.185\pm0.069$ & $0.182\pm0.073$
\\

EARL$^\dagger$ \cite{huang2023earl} & $0.347\pm0.102$ & $0.320\pm0.087$ & $0.270\pm0.065$ & $0.243\pm0.066$& $0.297\pm0.049$ & $0.260\pm0.048$
\\

GA-DDPG$^\dagger$~\cite{wang2022goal}  & $0.490\pm0.098$ & $0.459\pm0.109$ & $0.362\pm0.073$ & $0.341\pm0.071$& $0.382\pm0.059$ & $0.328\pm0.077$ 
\\

GAP-RL & $\textbf{0.580}\pm \textbf{0.037}$ & $\textbf{0.531}\pm \textbf{0.047}$ & $\textbf{0.510}\pm\textbf{0.041}$ & $\textbf{0.457}\pm\textbf{0.035}$ & $\textbf{0.487}\pm\textbf{0.042}$ & $\textbf{0.430}\pm\textbf{0.038}$
\\

\hline
\end{tabular}
$\dagger$: These methods are reproduced with their visual representations while keeping other settings consistent to make a fair comparison. \\
\end{threeparttable}
}
\vspace{-7mm}
\end{table*}

\begin{figure}[t]
\centering
    \includegraphics[width=8.5cm]{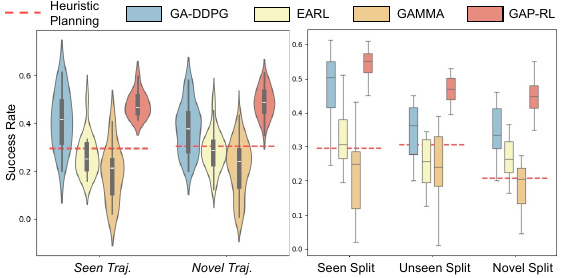}
    \caption{The visualized results of GAP-RL compared with other baselines, averaged across 5 random seeds.}
    \label{fig:main_exp} 
\vspace{-0.3cm} 
\end{figure}

During the training phase, all policies are trained under the Bezier setting, which encapsulates the most complex motion patterns for dynamic objects. To comprehensively evaluate the generalization capabilities across different dynamic settings, we evaluate all methods on the seen trajectory mode (Bezier setting) and three additional settings as novel trajectories, outlined in Table \ref{tab:main_exp} as \textit{Seen Traj.} and \textit{Novel Traj.}. We also investigate the methods' generalization abilities with respect to various object shapes. \textbf{Seen Split} and \textbf{Unseen Split}, derived from the Graspnet-1billion dataset \cite{fang2020graspnet}, consist of 12 and 10 objects respectively. \textbf{Novel Split} consists of 10 objects selected from the Acronym dataset \cite{eppner2021acronym}, representing novel items for the LoG model.

The evaluation results are displayed in Fig. \ref{fig:main_exp}, with detailed outcomes detailed in Table \ref{tab:main_exp}. In the \textit{Seen Traj.} setting, our GAP-RL framework outperforms the second-best method by approximately 18\% for seen objects and 34\% for novel objects. In the \textit{Novel Traj.} setting, GAP-RL achieves an improvement of over 15\% for seen objects and 32\% for novel objects compared to other methods. Compared with other visual presentations, GAP-RL with random Gaussian sampling demonstrates superior generalization across diverse object shapes and dynamic motion modes, likely due to its enhanced focus on grasp centers and hierarchical features from multiple grasps.

\subsection{Ablation Studies}

\begin{table}[t]
\centering
\caption{Ablations on Grasp Encoder. Averaged on all trajectories and objects over 5 random seeds.}
\label{tab:ablation}
\renewcommand\arraystretch{1.3}
\resizebox{0.45\textwidth}{!}{
\setlength{\tabcolsep}{1.2mm}
\small
\begin{tabular}{c|ccc}
\hline
\textbf{Methods} & $L=16$ & $L=32$ & $L=64$ 
\\ \hline
GP $\to$ KP & $0.348\pm0.072$ & $0.377\pm0.066$ & $0.408\pm0.088$
\\
GGN $\to$ PN & $0.423\pm0.051$ & $0.467\pm0.065$ & $0.459\pm0.064$ 
\\ 
GAP-RL & $\textbf{0.481}\pm\textbf{0.050}$ & $\textbf{0.502}\pm\textbf{0.054}$ & $\textbf{0.489}\pm\textbf{0.049}$
\\ \hline
\end{tabular}
}
\vspace{-0.5cm}
\end{table}

Table. \ref{tab:ablation} shows that replacing Gaussian Points (GP) with a rigid set of keypoints (KP) derived from the gripper model \cite{huang2023earl} significantly reduces model performance. Additionally, substituting the GraspGroupNet (GGN) with a PointNet (PN) architecture causes performance degradation, especially when the number of points is limited. Although this performance gap narrows as the number of points increases, computational efficiency significantly decreases. The results also demonstrate that GAP-RL maintains relative stability across different sampling numbers. These results highlight the vital roles that both random Gaussian points and the GGN play in the effectiveness of the Grasp Encoder, as discussed in Section \ref{GraspEncoder}. In the experiments, we sampled 16 Gaussian points to strike a balance between performance and computational efficiency.


\subsection{Qualitive Results}

\begin{figure}[t]
\centering
    \includegraphics[width=8cm]{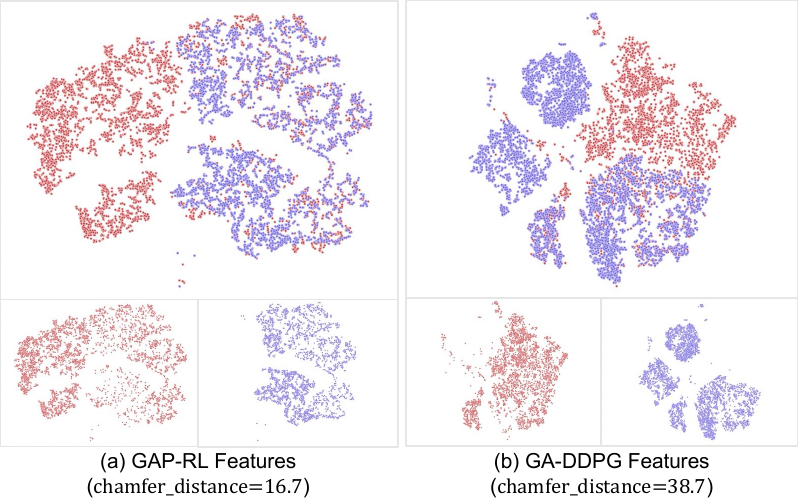}
    \caption{t-SNE visualization of geometry embeddings of different representations. Red: samples from perfect \textit{Synthetic} depth images using the simulation renderer. Blue: samples from \textit{Simulated} depth images using a stereo depth sensor \cite{zhang2023close}.
    \color{b}{The figure shows that GAP-RL features are indistinguishable between the two sample sets, indicating superior sim-to-real transferability of the Grasp Encoder.}}
    \label{fig:tSNE} 
\vspace{-0.2cm} 
\end{figure}

As shown in Fig. \ref{fig:tSNE}, we conduct a t-SNE visualization to compare the geometric features derived from the Grasp Encoder with those from the object points encoder of GA-DDPG \cite{wang2022goal}. In the visualization, red points represent features from perfect \textit{Synthetic} depth images produced by the simulation renderer, consistent with our training setups. Zhang et al. \cite{zhang2023close} designed a depth sensor simulator integrated into the Sapien simulator that effectively bridges the sim-to-real optical sensing gap. This simulator generates \textit{Simulated} depth images that closely mimic real-world conditions. We use this to re-evaluate GAP-RL and GA-DDPG, with features from these tests depicted as blue points in the figure.

\begin{table}[t]
\renewcommand{\arraystretch}{1.3}
\caption{Real robot experiments with 10 trails for each object.}
\vspace{-0.3cm}
\tabcolsep=0.05cm
\begin{center}
\resizebox{1.0\linewidth}{!}{
\begin{threeparttable}
\begin{tabular}{c|c|c c c c} 
\hline
\textbf{Method} & \textbf{ Average } & \textbf{C$^\dagger$ + T$^\dagger$} & \textbf{Conveyor} & \textbf{Turntable} & \textbf{Handover}  \\
\hline
Heuristic Planning & 42.9\% & 8 / 60 & 35 / 60  & 20 / 60 & 40 / 60 \\
GA-DDPG \cite{wang2022goal} & 57.9\% & 25 / 60 & 39 / 60  & 32 / 60 & 43 / 60 \\
GAP-RL & \textbf{76.3\%} & \textbf{39 / 60} & \textbf{49 / 60}  & \textbf{42 / 60} & \textbf{53 / 60} \\
\hline
\end{tabular}
$^\dagger$: C denotes Conveyor. T denotes Turntable.
\end{threeparttable}}
\vspace{-0.7cm}
\label{tab:real}
\end{center}
\end{table}

The results indicate that the features of GAP-RL, derived from \textit{Simulated} depth images, are indistinguishable from those from \textit{Synthetic} depth images. In contrast, the features from GA-DDPG are clearly distinct. This comparison underscores that Grasp Encoder has a superior expressive capability compared to the point features in GA-DDPG, significantly enhancing the sim-to-real transferability.


\subsection{Real Robot Experiments}

\begin{figure}[t]
\centering
    \includegraphics[width=8cm]{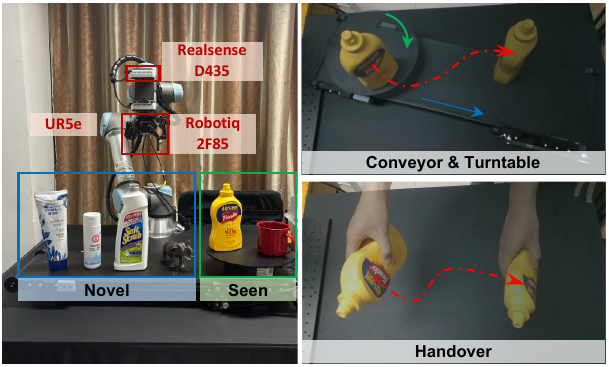}
    \caption{Real experimental settings: objects move through four settings including Conveyor, Turntable, Conveyor + Turntable and Handover. }
    \label{fig:realsettings} 
\vspace{-0.2cm} 
\end{figure}

Fig. \ref{fig:realsettings} illustrates the real-world experimental settings. We randomly sample two familiar objects from Seen Split and four novel objects for testing, conducting 10 trials for each object. The objects traverse the Conveyor, rotate on the Turntable, or navigate a combination of these mechanisms, mimicking the dynamic settings in the simulations. Additionally, the Handover setting tests the framework's adaptability to complex 3D dynamics.
Note that in real dynamic tasks, delays in observation and action execution are challenging to manage \cite{liotet2023delays} and can significantly impact the gripper's closing action as the robot approaches a moving object. To address this, following recent methodologies \cite{christen2023handover, wang2024genh2r}, we implement a sequential execution of three-step commands - \textit{Move Forward}, \textit{Grasp}, and \textit{Retract} once the agent is sufficiently close to the targeted grasps. Specifically, we compute the Euclidean distances $\{d_{t,j}\}_{j=1}^N$ and rotational distances $\{d_{r,j}\}_{j=1}^N$ as clarified in Equation \ref{eq.grasp_dist} for all grasps in each step. The RL policy continues executing until a grasp $\mathbf{\tilde{g}}$ satisfies $\tilde{d}_t<0.05 m$ and $\tilde{d}_r<0.05 rad$. Then $\mathbf{\tilde{g}}$ is set as the target, and we switch to sequential commands.

\begin{table}[t]
\renewcommand{\arraystretch}{1.3}
\caption{Results of Real Experiments. 10 Trails for 6 objects each.}
\label{tab:real_diffv}
\begin{center}
\begin{threeparttable}
\begin{tabular}{cc|cc} 
\hline
\multicolumn{1}{c|}{\textbf{Velocity}} & \textbf{Method} & \textbf{Conveyor} & \textbf{Turntable}\\
\hline
\multicolumn{1}{c|}{\multirow{2}{*}{Static}} & GAP-RL & \multicolumn{2}{c}{53 / 60} \\
\multicolumn{1}{c|}{} & \textit{w/o Explorer} & \multicolumn{2}{c}{48 / 60} \\
\hline
\multicolumn{1}{c|}{\multirow{2}{*}{Medium$^\dagger$}} & GAP-RL & 49 / 60  & 42 / 60 \\
\multicolumn{1}{c|}{} & \textit{w/o Explorer} & 40 / 60  & 36 / 60 \\
\hline
\multicolumn{1}{c|}{\multirow{2}{*}{High$^\ddagger$}} & GAP-RL & 34 / 60  & 32 / 60 \\
\multicolumn{1}{c|}{} & \textit{w/o Explorer} & 24 / 60  & 28 / 60  \\
\hline
\end{tabular}
$^\dagger$: $4\ cm/s$ for Conveyor and $6\ ^\circ/s$ for Turntable.\\
$^\ddagger$: $8\ cm/s$ for Conveyor and $12\ ^\circ/s$ for Turntable.
\end{threeparttable}
\vspace{-0.8cm}
\end{center}
\end{table}



To compare GAP-RL with other baselines, we selected GA-DDPG \cite{wang2022goal} as a representative learning-based method due to its strong performance in simulation, second only to our own. As depicted in Table \ref{tab:real}, GAP-RL outperforms these baselines, achieving an average success rate of 76.3\%, compared to 57.9\% for GA-DDPG and 42.9\% for Heuristic Planning. Notably, in the challenging Combined Conveyor and Turntable setting, our approach significantly outperforms the alternatives. 

As shown in Table \ref{tab:real_diffv}, GAP-RL's performance declines as the velocity of the objects increases, primarily due to the limitations in control frequency and the robotic arm's speed. Additionally, removing the Region Center Explorer results in a considerable drop in performance, underscoring its importance for sim-to-real transfer.

\section{LIMITATION}
Our methodology primarily relies on single-frame depth input and does not leverage temporal information for policy training. This limitation occasionally leads to failures due to collisions during grasping, where the object may move and slip out. 
The policy might be improved with a temporal-consistent representation by incorporating temporal association mechanisms as suggested in \cite{liu2023target, fang2023anygrasp}. Furthermore, the current framework is designed for single-object scenarios and cannot handle cluttered scenes or scenarios with external interference. This limitation could be addressed by integrating a segmentation module to isolate the target object.

\section{CONCLUSION}
We propose GAP-RL, a novel RL framework that transforms grasps into points for dynamic grasping tasks. By highlighting the significance of grasp poses in comparison to raw sensor inputs, our method offers a novel approach to dynamic grasping. We utilize Gaussian points to represent grasp poses and employ a hierarchical points encoder for efficient feature extraction, resulting in robust grasp features and better generalizability to novel objects. Comprehensive experiments conducted in both simulation and real-world scenarios demonstrate the effectiveness of the framework. Moving forward, we intend to explore the integration of temporal association information to further enrich grasp representation and optimize multi-frame policies.

{
\bibliographystyle{IEEEtran}
\bibliography{IEEEabrv, reference}
}

\addtolength{\textheight}{-12cm}   

\end{document}